\pdfoutput=1

\documentclass[11pt]{article}

\usepackage{acl}

\usepackage{times}
\usepackage{latexsym}

\usepackage[T1]{fontenc}

\usepackage[utf8]{inputenc}

\usepackage{microtype}

\usepackage{inconsolata}

\usepackage{graphicx}%
\usepackage{multirow}%
\usepackage{amsmath,amssymb,amsfonts}%
\usepackage{amsthm}%
\usepackage{mathrsfs}%
\usepackage[title]{appendix}%
\usepackage{booktabs}%
\usepackage{algorithm}%
\usepackage{algorithmicx}%
\usepackage{algpseudocode}%
\usepackage{listings}%

\usepackage{multirow}
\usepackage{multicol}
\usepackage{cleveref}
\usepackage{paralist}
\usepackage{xspace}
\usepackage[normalem]{ulem}
\usepackage{enumitem}

\newcommand{\mscript}[1]{\text{\scriptsize{#1}}}
\newcommand{\mbf}[1]{\boldsymbol{\mathbf{#1}}}

\newcommand{\ie}{\emph{i.e.,}\xspace}

\newcommand{\eg}{\emph{e.g.,}\xspace}

\crefname{equation}{Eq.}{Eq.}
\crefname{section}{Sec.}{Sec.}

\hypersetup{
    colorlinks=true,
    urlcolor=blue,
}

\title{
    A Survey on Neural Topic Models: Methods, Applications, and Challenges
}

\author{
    Xiaobao Wu$^\ddagger$ \qquad Thong Nguyen$^\dagger$ \qquad Anh Tuan Luu$^\ddagger$ \\
    $^\ddagger$Nanyang Technological University, \quad $^\dagger$National University of Singapore \\
    \texttt{xiaobao002@e.ntu.edu.sg} \quad \texttt{e0998147@u.nus.edu} \quad \texttt{anhtuan.luu@ntu.edu.sg}
}

\begin{document}
\maketitle
\begin{abstract}
    Topic models have been prevalent for decades to discover latent topics and infer topic proportions of documents in an unsupervised fashion.
    They have been widely used in various applications like text analysis and context recommendation.
    Recently, the rise of neural networks has facilitated the emergence of a new research field---Neural Topic Models (NTMs).
    Different from conventional topic models,
    NTMs directly optimize parameters without requiring model-specific derivations.
    This endows NTMs with better scalability and flexibility, resulting in significant research attention and plentiful new methods and applications.
    In this paper, we present a comprehensive survey on neural topic models concerning methods, applications, and challenges.
    Specifically, we systematically organize current NTM methods according to their network structures and introduce the NTMs for various scenarios like short texts and bilingual documents.
    We also discuss a wide range of popular applications built on NTMs.
    Finally, we highlight the challenges confronted by NTMs to inspire future research.
    We accompany this survey with a repository for easier access to the mentioned paper resources:
    \url{https://github.com/bobxwu/Paper-Neural-Topic-Models}.
\end{abstract}

\section{Introduction}
    Topic models seek to discover a set of latent topics from a collection of documents, depending on word co-occurrence information.
    Each topic represents an interpretable semantic concept and is described as a group of related words.
    For example, a topic about ``sports'' may relate to words like  ``baseball'', ``basketball'', and ``football''.
    Topic models also infer what topics a document contains (topic proportions) to reveal their underlying semantics.
    Due to their effectiveness and interpretability, 
    topic models have derived various downstream applications, such as document retrieval, content recommendation, opinion/event mining, and trend analysis \cite{blei2006dynamic,wang2011collaborative,boyd2017applications,duong2022,churchill2022evolution}.

\begin{figure*}[!t]
    \centering
    \includegraphics[width=\linewidth]{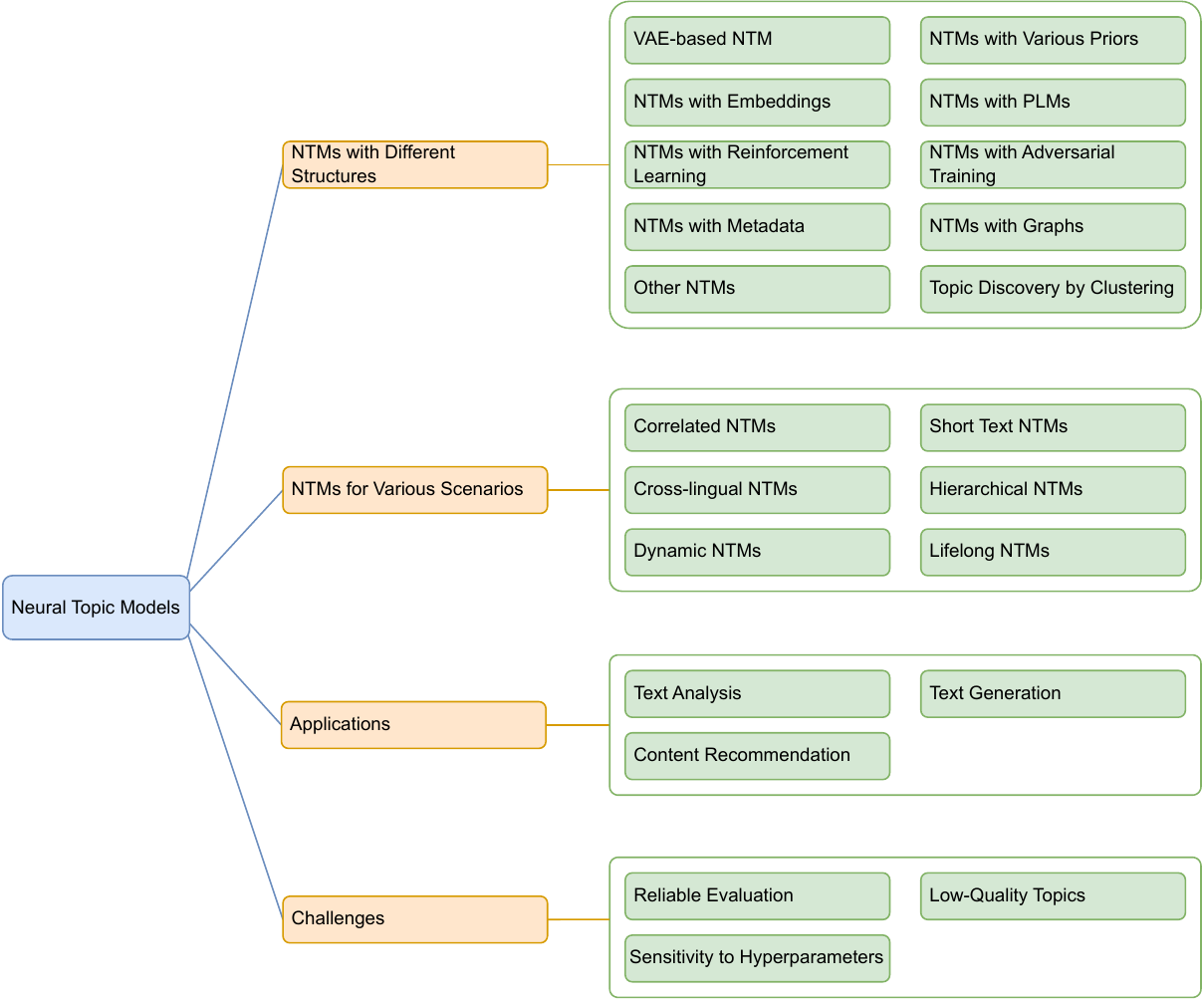}
    \caption{
        The overview of this survey: NTMs with different structures, NTMs for various scenarios, applications of NTMs, and challenges of NTMs.
    }
    \label{fig_overview}
\end{figure*}

    Conventional approaches to topic modeling embrace either probabilistic graphical models or non-negative matrix factorization.
    Approaches based on probabilistic graphical models, such as the classic Latent Dirichlet Allocation \citep[LDA, ][]{blei2003latent}, have been extensively explored for the past two decades \cite{blei2012probabilistic}.
    They mainly model the document generation process with topics as latent variables,
    and then infer model parameters through Variational Inference \cite{blei2017variational} or Monte Carlo Markov Chain (MCMC) methods like Gibbs sampling \cite{steyvers2007probabilistic}.
    Alternatively,
    another conventional topic model type uses non-negative matrix factorization.
    They directly discover topics by decomposing a term-document matrix into two low-rank factor matrices: one represents words and the other documents \cite{lee2000algorithms,kim2015simultaneous,shi2018short}.
    These conventional topic models have derived various model structures, such as supervised LDA \cite{mcauliffe2007supervised} and correlated LDA \cite{blei2006correlated}.
    Besides the basic topic modeling scenario,
    researchers have extended topic models to other diverse scenarios, \eg short text \cite{yan2013biterm,yin2014dirichlet}, cross-lingual \cite{mimno2009polylingual}, and dynamic topic modeling \cite{blei2006dynamic,wang2008continuous}.

    However, despite the achievements of these conventional methods,
    they generally confront two limitations:
    \begin{inparaenum}[(\bgroup\bfseries i\egroup)]
        \item
            \textbf{Inefficient and labor-intensive parameter inference}.
            These methods necessitate complicated model-specific derivations for parameter inference, and the inference complexity grows along with model complexity.
            Consequently, this requirement weakens their generalization ability to diverse model structures and application scenarios.
        \item
            \textbf{Limited scalability to large datasets}.
            Their inference algorithms typically are \emph{not} parallel, leading to significant time consumption.
            For example, training a probabilistic dynamic topic model DTM \citet{blei2006dynamic} on a dataset with 10k documents takes two days \cite{dieng2019dynamic}.
            Admittedly some parallel inference algorithms have been proposed \cite{newman2009distributed,wang2009plda,liu2011plda},
            but unfortunately they cannot straightforwardly fit other model structures and application scenarios.
    \end{inparaenum}
    As a result, how to design effective, flexible, efficient, and scalable topic models has become an urgent imperative in the research field.

\begin{figure*}[!t]
    \centering
    \includegraphics[width=0.8\linewidth]{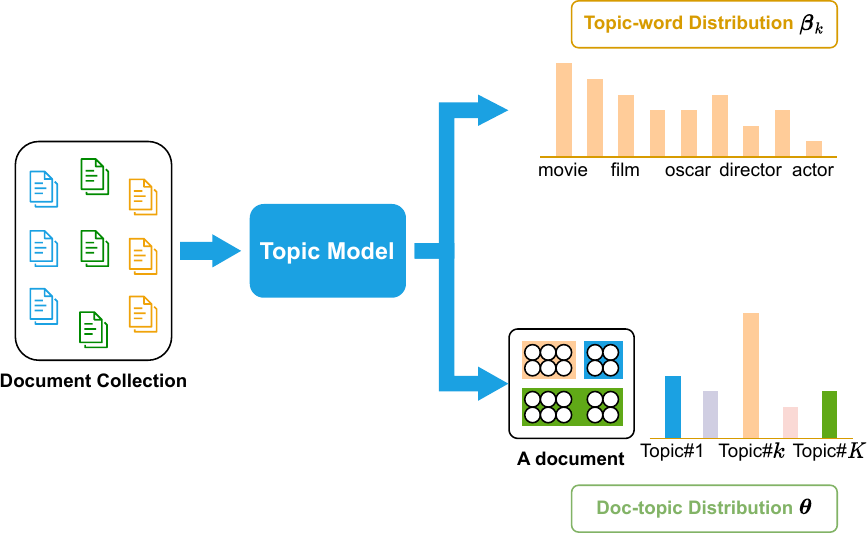}
    \caption{
        Illustration of topic modeling.
        Given a document collection,
        a topic model aims to discover $K$ latent topics, interpreted as distributions over words (topic-word distributions).
        It also infers topic proportions of each document (what topics a document contains), defined as distributions over all latent topics (doc-topic distributions).
        Here the topic-word distribution of Topic\#$k$, $\mbf{\beta}_{k}$, has related words like ``movie'', ``film'', and ``oscar'';
        the doc-topic distribution $\mbf{\theta}$ concentrates on Topic\#1 and Topic\#$k$.
    }
    \label{fig_topic_modeling}
\end{figure*}

    To overcome these challenges, Neural Topic Models (NTMs) have emerged as a promising solution.
    Unlike conventional topic models,
    NTMs can efficiently and flexibly infer model parameters through automatic gradient back-propagation
    by adopting deep neural networks to model latent topics, such as the popular Variational AutoEncoder \citep[VAE, ][]{Kingma2014autoencoding,Rezende2014}.
    This flexibility enables researchers to tailor model structures to fit diverse application scenarios.
    In addition, NTMs can seamlessly handle large-scale datasets by harnessing parallel computing facilities like GPUs.
    Owing to these advantages, NTMs have witnessed the exploration of numerous new methods and applications.

    Previously, \citet{zhao2021topic} provided a review with a primary focus on the methods of NTMs.
    However, their review is beset by the following limitations:
    \begin{inparaenum}[(\bgroup\bfseries i\egroup)]
        \item
            Their method taxonomy is incomplete because they ignore several recently proposed NTM methods,
            such as NTMs with contrastive learning, cross-lingual NTMs, and dynamic NTMs.
        \item
            They omit the popular applications based on NTMs, developed for a wide range of downstream tasks.
        \item
            They lack in-depth discussions on the challenges inherent in NTMs.
    \end{inparaenum}
    As a consequence, a more comprehensive review on NTMs is necessary for the research field.

    In this paper to address these limitations, we present an extensive and up-to-date survey of NTMs, which offers an in-depth and self-contained understanding of NTMs in terms of methods, applications, and challenges.
    We begin by systematically organizing existing NTMs according to their neural network structures, such as using embeddings or graph neural networks.
    We then introduce the NTMs designed for various prevalent topic modeling scenarios, \eg short text, cross-lingual, and dynamic topic modeling, covering a wider range than the early survey \cite{zhao2021topic}.
    Moreover while omitted by the previous survey \cite{zhao2021topic}, we also organize and discuss the popular applications based on NTMs, developed for diverse tasks like text analysis and text generation.
    Finally we summarize the key challenges for NTMs in detail to motivate future research directions.
    \Cref{fig_overview} depicts the overview of our survey.
    We conclude the main contributions of this paper as follows:
    \begin{itemize}[leftmargin=*]
        \item
            We extensively review methods of neural topic models through detailed discussions and comparisons, covering variants with different network structures.
        \item
            We include a broader range of popular topic modeling scenarios and provide detailed background information for each scenario,
            accompanied by easy-to-understand illustrations and related neural topic models.
        \item
            We introduce popular applications based on neural topic models, developed to tackle various tasks such as text analysis and generation.
        \item
            We highlight the current vital challenges faced by neural topic models in detail to facilitate future research;
            Motivated by this, we propose a new topic diversity metric that measures diversity along with word semantics, which more agrees with human judgment.
    \end{itemize}

\section{Preliminary of Topic Models} \label{sec_preliminary}
    In this section,
    we introduce the preliminary of topic modeling, including the problem setting, notations, and evaluation.
    Then we present the most basic and popular NTM in the framework of Variational AutoEncoder (VAE).

    \subsection{Problem Setting and Notations}
        We introduce the problem setting and notations of topic modeling following LDA \cite{blei2003latent}.
        Consider a collection of $N$ documents
        with $V$ unique words (vocabulary size), and a document is denoted as $\mbf{x}$.
        As illustrated in \Cref{fig_topic_modeling},
        topic models aim to discover $K$ latent topics from this collection.
        The number of topics $K$ is a hyperparameter, usually determined by researchers manually according to the characteristics of datasets and their target tasks.
        Each topic is defined as a distribution over the vocabulary, \ie \textbf{topic-word distribution},
        $\mbf{\beta}_{k} \in \mathbb{R}^{V}$.
        Then the topic-word distribution matrix of all topics is
        \begin{equation}
            \mbf{\beta}=(\mbf{\beta}_{1},\dots,\mbf{\beta}_{K}) \in \mathbb{R}^{V \times K} .
        \end{equation}
        In addition, topic models also infer the topic distribution of a document (\textbf{doc-topic distribution}): $\mbf{\theta} \in \Delta_{K}$, implying what topics a document contains.
        Here $\theta_{k}$ refers to the proportion of Topic\#$k$ in the document, and $\Delta_{K}$ denotes a probability simplex $\Delta_{K} = \{ \mbf{\theta} \in \mathbb{R}^{K}_{+} | \sum_{k=1}^{K} \theta_{k} = 1 \}$.

    \subsection{Evaluation of Topic Models} \label{sec_preliminary_evaluation}
        Given the absence of ground-truth labels in topic modeling tasks,
        how to reliably and comprehensively evaluate topic models remains inconclusive in the research community.
        We introduce currently the most prevalent evaluation methods employed for assessing topic models as follows.

        \subsubsection{Perplexity}
            Perplexity, borrowed from language models, measures how a model can predict new documents.
            It is measured as the normalized log-likelihood of held-out test documents.
            Perplexity has been used for years to evaluate topic models.
            Nevertheless, prior studies have empirically demonstrated that perplexity inaccurately reflects the quality of discovered topics as it often contradicts human judgment \cite{chang2009reading}.
            Furthermore, computing log-likelihood is inconsistent among different topic models.
            This is because they apply various sampling or approximation techniques \cite{wallach2009evaluation,buntine2009estimating} as well as diverse modeling approaches for topic-word distributions and doc-topic distributions.
            For instance, certain methods normalize topic-word distributions with respect to topics, some with respect to words, and others opt to keep them unnormalized.
            These disparities bring challenges to equitable comparisons.
            Finally, perplexity may not evaluate the practical utility of topic models since users typically employ topic models for content analysis rather than generating new documents \cite{zhao2021topic,hoyle2022neural}.
            Due to these reasons, perplexity has waned in popularity for topic model evaluation in the recent research field.

        \subsubsection{Topic Coherence}
            Rather than predictive abilities, researchers turn to evaluating the quality of produced topics.
            For this purpose, researchers propose topic coherence to measure the coherence among the most related words of topics, \ie top words (determined by topic-word probabilities).
            Experiments showcase that topic coherence can agree with the human evaluation on topic interpretability \cite{lau2014machine}.
            For example, one widely-used coherence metric is Normalized Point-wise Mutual Information \citep[NPMI, ][]{bouma2009normalized,Newman2010,lau2014machine}.~\footnote{\url{https://github.com/jhlau/topic_interpretability}}
            Specifically, the NPMI score between two words $(x_i, x_j)$ is calculated as follows:
            \begin{align}
                \mathrm{NPMI}(x_i, x_j) = \frac{\log \frac{p(x_i, x_j) + \epsilon}{p(x_i) p(x_j)}}  {-\log p(x_i, x_j)+\epsilon}.
                \label{eq_NPMI}
            \end{align}
            It computes the normalized mutual information of two words,
            and then takes the average of all word pairs in all topics.
            Here $\epsilon$ is to avoid zero; $p(x_i)$ is the probability of word $x_i$, and $p(x_i, x_j)$ is the co-occurrence probability of $(x_i, x_j)$.
            These probabilities are estimated as their occurrence frequencies in a reference corpus.
            The reference corpus can be either internal (the training set) or external (\eg Wikipedia articles).
            Basically, a large external corpus is recommended because it can alleviate the influence of data bias in training sets and facilitate fair topic coherence comparisons across different datasets.
            
            Later, \citet{roder2015exploring} propose a new metric, $C_V$, which calculates the cosine similarity between NPMI score vectors \cite{krasnashchok2018improving}.
            Given the top $T$ words of a topic, $(x_1, x_2, \dots, x_T)$,
            the exact calculation of $C_V$ is formulated as
            \begin{align}
                &C_V \!\!=\!\! \frac{1}{T} \sum_{i=1}^{T} \cos( \mbf{v}_{\mscript{NPMI}}(x_i), \mbf{v}_{\mscript{NPMI}}(\{x_i\}_{i=1}^{T}) ) \\
                &\mbf{v}_{\mscript{NPMI}}\left(x_i\right) \!=\! \left\{\mathrm{NPMI}\left(x_i, x_j\right)\right\}_{j=1, \ldots, T} \\
                &\mbf{v}_{\mscript{NPMI}}\left(\{x_i\}_{i=1}^{T}\right) \!\!=\!\! \left\{\sum_{i=1}^T \mathrm{NPMI}\left(x_i, x_j\right)\right\}_{j=1, \ldots, T}.
            \end{align}
            The NPMI score computation follows \Cref{eq_NPMI}.
            \citet{roder2015exploring} empirically demonstrate that $C_V$ outperforms previous coherence metrics, NPMI, UCI, and UMass \cite{mimno2011optimizing}, since $C_V$ is more consistent with human judgment (See \citet{roder2015exploring} for experimental results).

            We would like to recommend the Palmetto tool~\footnote{\url{https://github.com/dice-group/Palmetto}} to compute topic coherence.
            It includes almost all common coherence metrics and provides a pre-processed Wikipedia article collection as the reference corpus for easier reproducibility.

        \subsubsection{Topic Diversity}

            To further evaluate the quality of topics, topic diversity is introduced to measure the difference between topics.
            This is driven by the anticipation that topics should exhibit diversity rather than redundancy
            thereby enabling the comprehensive disclosure of latent semantics in corpora.
            At present, researchers propose the following common diversity metrics:
            \begin{itemize}[leftmargin=*,itemsep=0pt]
                \item
                    \citet{Nan2019} propose Topic Uniqueness (TU) which computes the average reciprocal of top word occurrences in topics.
                    In detail given $K$ topics and the top $T$ words of each topic,
                    TU is computed as
                    \begin{equation}
                        \mathrm{TU} = \frac{1}{K} \sum_{k=1}^{K} \frac{1}{T} \sum_{x_i \in t(k)} \frac{1}{\#(x_{i})} \label{eq_TU}
                    \end{equation}
                    where $t(k)$ means the top word set of the $k$-th topic, and $\#(x_i)$ denotes the occurrence of word $x_i$ in the top $T$ words of all topics.
                    TU ranges from $1/K$ to $1.0$, and a higher TU score indicates more diverse topics.
                \item
                    \citet{burkhardt2019decoupling} propose Topic Redundancy (TR) that calculates the average occurrences of a top word in other topics.
                    Its computation is
                    \begin{equation}
                        \mathrm{TR} = \frac{1}{K} \sum_{k=1}^{K} \frac{1}{T} \sum_{x_i \in t(k)} \frac{\#(x_i) - 1}{K - 1}. \label{eq_TR}
                    \end{equation}
                    A lower TR score means more diverse topics.
                \item
                    \citet{dieng2020topic} propose topic diversity (TD) which computes the proportion of unique top words of topics:
                    \begin{equation}
                        \mathrm{TD} = \frac{1}{K} \sum_{k=1}^{K} \frac{1}{T} \sum_{x_i \in t(k)} \mathbb{I}(\#(x_i)) \label{eq_TD}
                    \end{equation}
                    where $\mathbb{I}(\cdot)$ is a indicator function that equals 1 if $\#(x_i) = 1$ and equals 0 otherwise.
                    TD ranges from $0$ to $1.0$, and a higher TD score indicates more diverse topics.
            \end{itemize}

            These metrics all measure topic diversity based on the uniqueness of individual words.
            They posit that diversity is optimal when all topics are characterized by distinct top words.
            However, we question these diversity metrics because certain topics naturally share the same words.
            For example, the word ``chip'' could be shared by the topics of ``potato chip'' and ``electronic chip'';
            Similarly, the word ``apple'' may be covered by the topics of ``fruit'' and ``company''.
            This issue remains unresolved for reliable diversity evaluation.
            We in this paper propose a new diversity metric to address this issue (See details in \Cref{sec_TSD}).

\begin{figure}
    \centering
    \includegraphics[width=\linewidth]{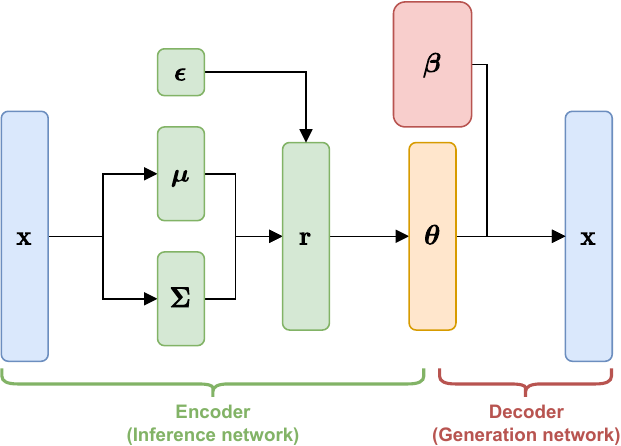}
    \caption{
        Illustration of a VAE-based NTM.
        It mainly contains an encoder (inference network) and a decoder (generation network).
        The encoder outputs doc-topic distribution $\mbf{\theta}$ from input document $\mbf{x}$ through MLPs using the reparameterization trick where $\mbf{\epsilon} \sim \mathcal{N}(\mbf{0}, \mbf{I})$.
        The decoder reconstructs the input document from $\mbf{\theta}$ with $\mbf{\beta}$ as the topic-word distribution matrix.
        The objective includes reconstruction error and KL divergence.
    }
    \label{fig_VAE-based_NTM}
\end{figure}

        \subsubsection{Downstream Task Performance}
            Except for the coherence and diversity to measure topic quality,
            researchers also resort to extrinsic performance:
            they use doc-topic distributions $\mbf{\theta}$ as low-dimensional document features and evaluate their quality on downstream tasks.
            These tasks mainly consist of document classification and document clustering.
            For document classification,
            researchers train ordinary classifiers (\eg SVMs or Random Forests) with learned doc-topic distributions as document features and then predict the labels of testing documents.

            The performance can be evaluated by accuracy or F1 scores.
            For document clustering, the common way is to use the most significant topic in a doc-topic distribution as the clustering assignment of a document.
            Another way is to apply clustering algorithms, \eg K-Means or DBSCAN, on doc-topic distributions \cite{zhao2021neural}.
            The clustering performance can be measured by Purity and Normalized Mutual Information \citep[NMI, ][]{Manning2008}.

        \subsubsection{Visualization}
            Finally, researchers visualize topic models for evaluation.
            The typical visualization method is to show the top words of topics and doc-topic distributions, such as using pyLDAvis~\footnote{\url{https://github.com/bmabey/pyLDAvis}} \cite{sievert2014ldavis} or word cloud~\footnote{\url{https://github.com/amueller/word_cloud}}.
            Another strategy is to cluster documents on a 2D canvas by reducing the dimension of doc-topic distributions with tools like t-SNE \cite{Maaten2008}.

    \subsection{Basic NTM based on VAE}
        We introduce the most basic and popular NTM based on the Variational AutoEncoder (VAE) framework with the neural variational inference technique \cite{Miao2016,Srivastava2017}.
        As illustrated in \Cref{fig_VAE-based_NTM},
        a VAE-based NTM mainly contains an encoder (inference network) and a decoder (generation network).
        The encoder is to infer doc-topic distributions from documents.
        To be specific,
        we use a latent variable $\mbf{r} \in \mathbb{R}^{K} $ following a logistic normal prior
        \begin{align}
            p(\mbf{r}) = \mathcal{LN}(\mbf{\mu}_{0}, \mbf{\Sigma}_{0})            
        \end{align}
        where $\mbf{\mu}_{0}$ and $\mbf{\Sigma}_{0}$ are the mean vector and diagonal covariance matrix respectively.
        Here the prior distribution is specified with Laplace approximation \cite{Hennig2012} to approximate a symmetric Dirichlet prior as
        $\mu_{0,k} = 0$ and $\Sigma_{0, kk} = (K-1) / \alpha K$ with hyperparameter $\alpha$ \cite{Srivastava2017}.
        The variational distribution is modeled by parameters $\Theta$ as
        \begin{align}
            q_{\Theta}(\mbf{r} | \mbf{x}) = \mathcal{N}(\mbf{\mu}, \mbf{\Sigma}) .
        \end{align}
        We compute $\mbf{\mu}$ and $\mbf{\Sigma}$ with encoder networks $f_{\Theta_{1}}$ and $f_{\Theta_{2}}$:
        \begin{align}
            \mbf{\mu} &= f_{\Theta_{1}}(\mbf{x}) \\
            \mbf{\Sigma} &= \mathrm{diag}(f_{\Theta_{2}}(\mbf{x}))
        \end{align}
        where $\Theta = \{ \Theta_{1}, \Theta_{2} \}$ and $\mathrm{diag}(\cdot)$ denotes transforming a vector to a diagonal matrix.
        In practice, we employ MLPs as the encoder networks and transform document $\mbf{x}$ into a Bag-of-Words (BoW) vector as inputs.
        Then to avoid gradient variance \cite{Kingma2014autoencoding,Rezende2014},
        we sample $\mbf{r}$ through the reparameterization trick by sampling a random variable $\mbf{\epsilon}$:
        \begin{align}
            \mbf{r} = \mbf{\mu} + (\mbf{\Sigma})^{1/2} \mbf{\epsilon} \quad \text{where} \quad \mbf{\epsilon} \sim \mathcal{N}(\mbf{0}, \mbf{I}).
        \end{align}
        We model the doc-topic distribution $\mbf{\theta}$ with a softmax function to restrict it on a simplex:
        \begin{align}
            \mbf{\theta} = \mathrm{softmax}(\mbf{r}) .
        \end{align}
        The decoder is to generate documents from doc-topic distributions.
        Specifically,
        we use a decoder network parameterized by $\Phi$: $f_{\Phi}(\mbf{\theta}) = \mathrm{softmax}(\mbf{\beta}\mbf{\theta})$ which represents the generation probability of each word.
        Here $\Phi = \{ \mbf{\beta} \}$.
        Then we sample words from its multinomial distribution: $ x \sim \mathrm{Mult}(f_{\Phi}(\mbf{\theta}))$.
        Following the Evidence Lower BOund (ELBO) of VAE, we formulate the learning objective of NTM as
        \begin{align}
            \min_{\Theta,\Phi} -\mathbb{E}_{q_{\Theta}(\mbf{\theta}|\mbf{x})} \left[ \log p_{\Phi}(\mbf{x}|\mbf{\theta}) ] + \mathrm{KL} [ q_{\Theta}(\mbf{r}|\mbf{x}) \| p(\mbf{r}) \right].
        \end{align}
        The first term is the negative expected log-likelihood, \emph{i.e.}, the reconstruction error, where $p_{\Phi}(\mbf{x}|\mbf{\theta})$ denotes the generation probability of $\mbf{x}$.
        As we sample words from the multinomial distribution,
        the first term becomes $-\mbf{x}^{\top} \log (f_{\Phi}(\mbf{\theta})) $.
        The second term is the Kullback-Leibler (KL) divergence between the variational and prior distributions, which can be computed through an analytical form \cite{Srivastava2017}.
        It is also known as a regularization term.

        The above is the fundamental structure of a VAE-based NTM, followed by pioneer studies like NVDM \citet{Miao2016} and ProdLDA \citet{Srivastava2017}.
        Based on this,
        NTMs with different structures are proposed to further improve performance and deal with various application scenarios.

\section{NTMs with Different Structures}

    Apart from the basic VAE structure mentioned in \Cref{sec_preliminary},
    we in this section introduce NTMs with more different structures.

        \subsection{NTMs with Various Priors}
            VAE-based NTMs commonly employ Gaussian (Normal) as priors since it is easy to apply the reparameterization trick and compute the analytical form of KL divergence.
            Besides Gaussian priors, NTMs also leverage other various priors.
            \citet{Miao2017} propose new priors like Gaussian softmax and the stick-breaking process.
            \citet{zhang2018whai} use a Weibull distribution to approximate gamma distributions.
            \citet{joo2020dirichlet} leverage an auxiliary uniform distribution to approximate the cumulative distribution function of gamma.
            As Dirichlet priors are important for topic modeling,
            \citet{burkhardt2019decoupling} utilize the proposal function of a rejection sampler for a gamma distribution to approximate Dirichlet priors.
            \citet{tian2020learning} draw from the rounded posterior distribution to approximate Dirichlet samples.

        \subsection{NTMs with Embeddings}
            Alternative to directly modeling topics,
            \citet{Miao2017} propose to decompose topics as two embedding parameters:
            \begin{equation}
                \mbf{\beta} = \mbf{W}^{\top}\mbf{T}. \label{eq_beta_matmul}
            \end{equation}
            Here $\mbf{W} \in \mathbb{R}^{D \times V}$ denotes $V$ word embeddings,
            and $\mbf{T} \in \mathbb{R}^{D \times K}$ denotes $K$ topic embeddings,
            where $D$ is the dimension of embedding space.
            Then \citet{dieng2020topic} follow this setting and propose ETM (Embedding Topic Model).
            ETM facilitates topic learning by initializing $\mbf{W}$ with pre-trained word embeddings like Word2Vec \cite{Mikolov2013} or GloVe \cite{pennington2014glove}.
            This approach also confers flexibility and efficiency to other topic modeling scenarios.
            For instance, it is much cheaper to repeat topic embeddings for each time slice in dynamic topic modeling than repeating the entire topic-word distribution matrix \cite{dieng2019dynamic}.

            Alternatively,
            \citet{zhao2021neural} propose NSTM.
            It also models topics as embeddings, but uses the optimal transport distance between doc-topic distributions and input documents to measure the reconstruction error.
            \citet{wang2022representing} share the same idea and instead use conditional transport distance.
            \citet{duan2022bayesian} learn a group of global topic embeddings for task-specific adaptations.
            \citet{xu2022hyperminer} propose HyperMiner, using hyperbolic embeddings to model topics.
            Due to the tree-likeness property of hyperbolic space, they can capture the hierarchy among topics.

            Differently, \citet{wu2023effective} propose ECRTM, which models the topic-word distribution matrix as
            \begin{equation}
                \beta_{jk} = \frac{\exp(-\| \mbf{w}_{j} - \mbf{t}_{k} \|^{2} / \tau)}{ \sum_{k'=1}^{K} \exp(-\| \mbf{w}_{j} - \mbf{t}_{k'}\|^{2} / \tau)}.
            \end{equation}
            Here $\beta_{jk}$ denotes the correlation between $j$-th word and $k$-th topic with $\tau$ as a temperature hyperparameter; $\mbf{w}_{j}$ is the $j$-th word embedding in $\mbf{W}$, and $\mbf{t}_{k}$ is the $k$-th topic embedding in $\mbf{T}$.
            It computes the Euclidean distance between topic and word embeddings and normalizes overall topics in a softmax manner.
            This works together with a clustering regularization method.
            The regularization considers topic embeddings as cluster centers and word embeddings as cluster samples;
            then it forces topic embeddings to be the centers of separately aggregated word embeddings by optimal transport.
            This effectively avoids the topic collapsing issue where topics are repetitive to each other.

        \subsection{NTMs with Metadata}
            While common NTMs learn topics in an unsupervised manner (only using documents),
            NTMs can also leverage the metadata of documents to guide topic modeling,
            similar to supervised LDA \cite{mcauliffe2007supervised}.
            In detail, \citet{Card2018a} introduce \textsc{Scholar}, a NTM that can incorporate various metadata of documents.
            It encodes a document together with its labels (\emph{e.g.}, sentiment) and covariates (\emph{e.g.}, publication year),
            and generates the document conditioned on the covariates.
            \citet{korshunova2019discriminative} model the generation of documents and labels together in a discriminative way;
            then train their model with mean-field variational inference.
            They can also incorporate a variety of data modalities like images.
            \citet{wang2020attention} jointly model topics and train a RNN classifier to predict document labels.
            They are connected by an attention mechanism.
            \citet{wang2021layer} incorporate document networks in a NTM and jointly reconstruct documents and networks.

        \subsection{NTMs with Graph Neural Networks}
            In addition to traditional BoW (Bag-of-Words) as inputs, several NTMs use graph neural networks to model documents.
            Specifically, \citet{zhu2018} transform documents into biterm graphs and follow the VAE framework to reconstruct the input graphs.
            A biterm refers to an unordered word pair that co-occurred in the same document, originally from \citet{yan2013biterm}.
            Similarly, \citet{yang2020graph,zhou2020neural} use a bipartite graph of documents and words, connected by word occurrences or TF-IDF values.
            \citet{wang2021extracting} use word co-occurrence and semantic correlation graphs.
            \citet{wang2022topic} focus on graph topic modeling with micro-blogs.
            \citet{zhu2023graph} propose a graph neural topic model to incorporate commonsense knowledge.

        \subsection{NTMs with Adversarial Training}
            Some studies focus on employing
            adversarial training to facilitate topic modeling.
            \citet{wang2019atm} follows the idea of Generative Adversarial Network (GAN):
            they use a generator to generate ``fake'' documents from a random Dirichlet sample and then use a discriminator to distinguish the generated documents from real ones.
            Note that their model cannot infer doc-topic distributions because it directly maps documents to representations based on TF-IDF.
            To lift this limitation,
            \citet{wang2020neural} propose to use bidirectional adversarial training, which can meanwhile infer doc-topic distributions.
            \citet{hu2020neural} further present an extension that uses two cycle-consistency constraints to generate informative representations.

        \subsection{NTMs with Pre-trained Language Models}
            Researchers frequently combine NTMs with pre-trained language models.
            Pre-trained language models based on Transformers \cite{vaswani2017attention} have been prevalent in NLP fields, which are pre-trained on large-scale corpora to capture contextual linguistic features.
            Multiple studies leverage contextual features from these pre-trained models to provide richer information than conventional BoW.
            For instance, \citet{bianchi2021pre} input the concatenation of BoW and the contextual document embeddings from Sentence-BERT \cite{reimers2019sentence},
            and then reconstruct BoW as previous work.
            \citet{hoyle2020improving} propose to distill knowledge from BERT \cite{devlin2018bert} to NTMs.
            In detail,
            they produce pseudo BoW from the predictive word probability of BERT.
            Then their NTM reconstructs both the real and pseudo BoW.
            \citet{bianchi2020cross,mueller2021fine} employ multilingual BERT to infer cross-lingual doc-topic distributions for zero-shot learning but they cannot discover aligned cross-lingual topics.

\begin{figure*}[!t]
    \centering
    \includegraphics[width=0.7\linewidth]{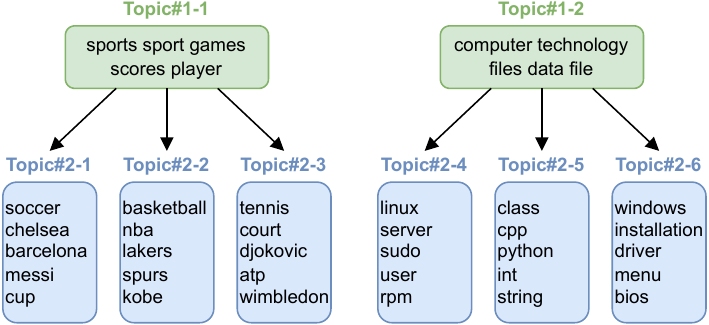}
    \caption{
        Illustration of hierarchical topic modeling.
        Topics at each level cover different semantic granularity: child topics are more specific to parent topics.
        Topic\#2-1 denotes the first topic of the second layer in the topic structure.
    }
    \label{fig_hierarchical_topic_modeling}
\end{figure*}

        \subsection{NTMs with Contrastive Learning}
            As a self-supervised learning fashion,
            contrastive learning has been employed to facilitate NTMs \cite{hadsell2006dimensionality,nguyen2022adaptive,nguyen2024kdmcse}.
            The idea of contrastive learning is to measure the similarity relations among sample pairs in a representation space \cite{van2018representation}.
            \citet{nguyen2021contrastive} propose the contrastive learning on doc-topic distributions
            where they build positive and negative pairs by sampling salient words of documents.
            Differently, \citet{wu2022mitigating} directly sample positive and negative pairs based on the topic semantics of documents to capture relations among samples.
            Specifically, they quantize doc-topic distributions following \citet{Wu2020short} and then sample documents with the same quantization indices as positive pairs and different indices as negative pairs.
            Their method can also capture the similarity relations among additional augmented data.
            \citet{zhou2023improving} improve topic disentanglement with contrastive learning on word and topic embeddings.
            \citet{han2023unified} cluster documents, compute term weights, and make NTMs reconstruct salient words.
            They also use contrastive learning to refine doc-topic distributions where positive samples come from pre-trained language models.
            Besides document-level contrastive learning,
            \citet{nguyen2024topic} also consider topic-level and propose a multi-objective contrastive learning method.

        \subsection{NTMs with Reinforcement Learning}
            Reinforcement learning has been utilized to guide the learning process of NTMs.
            To be specific, \citet{gui2019neural} enhance NTMs with a reinforcement learning framework.
            They evaluate topic coherence performance during training and use this performance as reward signals to guide the learning of topic modeling.
            \citet{costello2023reinforcement} follow this idea and add more improvements like using sentence embeddings, adding a weighting term to the ELBO, and tracking topic diversity and coherence during training.

        \subsection{Other NTMs}
            Apart from the aforementioned methods, we introduce NTMs with other structures.

            Before the invention of VAE-based NTMs,
            researchers have different attempts to model latent topics with neural networks.
            Some studies focus on NTMs in the autoregressive framework.
            \citet{larochelle2012neural} propose an autoregressive NTM, called DocNADE.
            Inspired by Replicated Softmax \cite{hinton2009replicated},
            DocNADE predicts the probability of a word in a document conditioned on its hidden state which is conditioned on previous words.
            Then it interprets topics with a hidden state and infers doc-topic distributions with the hidden states of the document.
            \citet{gupta2019document} extend DocNADE by modeling the bi-directional dependencies between words.
            \citet{gupta2019texttovec} then use a LSTM to enable DocNADE to incorporate external knowledge.

            \citet{cao2015novel} also propose an early NTM before VAE-based NTMs.
            Their approach predicts how an n-gram correlates with documents.
            It computes the representation of an n-gram by transforming the accumulation of the word embeddings from Word2Vec \cite{Mikolov2013} and projects documents into representations with a look-up matrix table.
            In this way,
            they model topic-word distributions as the n-gram representations
            and model doc-topic distributions as the projected document representations.
            For training,
            it uses the document of the n-gram as a positive and randomly samples documents that do \emph{not} contain this n-gram as negatives.

            \citet{lin2019sparsemax} replace the softmax function with the sparsemax to enhance the sparsity in doc-topic distributions.
            \citet{Nan2019} use Wasserstein AutoEncoder (WAE) to model topics,
            which minimizes the Wasserstein distance between generated documents and input documents.
            \citet{rezaee2020discrete} propose a NTM without using the reparameterization trick.
            They generate discrete topic assignments from RNN inspired by \citet{dieng2017topicrnn}.
            \citet{wu2021discovering} focus on discovering latent topics from long-tailed corpora.
            They propose a causal inference framework to analyze how the long-tailed bias influences topic modeling.
            Then they use a simple but effective casual intervention method to mitigate such influence.
    
            \citet{pham2024topicgpt} propose TopicGPT, prompting Large Language Models (LLMs) to augment seed topics.
            It defines each topic as a textual description, rather than the word distributions in LDA.
            It also prompts LLMs to assign a single or more topics to a document, like a classification task.
            We emphasize that TopicGPT cannot give the topic-word distributions or doc-topic distributions for downstream applications.

    \subsection{Topic Discovery by Clustering}
        We discuss a special type of approach that discovers latent topics by clustering instead of modeling the generation process of documents.
        They typically leverage traditional word embeddings such as Word2Vec \cite{Mikolov2013} or contextual embeddings from pre-trained language models.
        \textbf{We must clarify that some of these methods differ from the aforementioned ordinary topic models}.
        This is because they can only produce topics but cannot infer doc-topic distributions as required.
        Accordingly, their one advantage is their enhanced computational efficiency.
        In detail, \citet{thompson2020topic} straightforwardly cluster token-level word embeddings from pre-trained models like BERT and GPT-2
        and produce topics from the words assigned to clusters.
        Similarly, \citet{Sia2020,zhang2022neural} cluster word embeddings and interpret hidden topics by sampling words from clusters via term weights like TF-IDF.
        \citet{angelov2020top2vec} propose Top2Vec.
        It obtains document and word embeddings through Doc2Vec \cite{le2014doc2vec}.
        Then it reduces the dimension of document embeddings with UMAP and clusters them with HDBSCAN.
        It extracts the n-closest words of a cluster to represent a topic. 
        Following Top2Vec, \citet{grootendorst2022bertopic} propose BERTopic.
        BERTopic extracts words of a clustering based on c-TF-IDF, which calculated the TF-IDF of a word over a cluster.
        It estimates the doc-topic distribution based on the term weights within a given document.

    \subsection{NTM with Semantic-Relations}
        \citet{wu2024fastopic} propose FASTopic following a new paradigm rather than VAE-based or clustering-based ones.
        Similar to Top2Vec and BERTopic,
        FASTopic leverages document embeddings, like from pretrained Transformers \cite{reimers2019sentence,pan2023fact,wu2024updating}.
        Differently, it conducts optimal transport \cite{peyre2019computational} between document and topic embeddings and uses the transport plans to model doc-topic distributions.
        In the same way,
        it models topic-word distributions with the transport plans between topic and word embeddings.
        Then it optimizes topic and word embeddings by reconstruction with these semantic relations.
        This avoids the previous complicated VAE or simple clustering structures,
        leading to a neat and efficient paradigm.

    \section{NTMs for Various Scenarios}
        Apart from the basic scenario on normal documents,
        we in this section introduce NTMs tailored for various use case scenarios, such as hierarchical, cross-lingual, and dynamic topic modeling.
        We present the background of each scenario and its related NTMs.

        \subsection{Hierarchical NTMs}
            Similar to conventional topic models \cite{griffiths2003hierarchical,teh2004sharing,blei2010nested},
            NTMs can discover hierarchical topics to reveal topic structures from general to specific.
            Topics at each level in a hierarchy cover different semantic granularity:
            child topics tend to be more specific to their parent topics.
            As shown in \Cref{fig_hierarchical_topic_modeling},
            a topic about ``sports'' can derive more specific child topics, like ``soccer'', ``basketball'', and ``tennis'';
            a topic about ``computer'' also has specific child topics like ``linux'', ``programming'', and ``windows''.
            In addition, hierarchical topic modeling can relieve the challenge of determining the number of topics to some extent \cite{blei2010nested}.

            To discover hierarchical topics, some NTMs follow the previous non-parametric setting which allows topic hierarchies to grow dynamically.
            \citet{isonuma2020tree} propose a tree-structured neural topic model
            with two doubly-recurrent neural networks over the ancestors and siblings respectively \cite{alvarez2017tree}.
            Note that the tree structure is unbounded, \ie it can be dynamically updated in a heuristic way during training.
            \citet{pham2021neural} follow this spirit and jointly handle hierarchical topics and document visualization.
            \citet{chen2021tree} leverage a stick-breaking process as prior for non-parametric modeling.

            Later, the parametric fashion has attracted more attention, which presets the width and height of a topic hierarchy before learning.
            This leads to the main issue that topic hierarchies cannot grow dynamically.
            \citet{chen2021hierarchical} propose manifold regularization on topic hierarchy learning.
            \citet{duan2021sawtooth} propose a Sawtooth Connection to model topic dependencies across hierarchical levels based on the model structure of ETM \cite{dieng2020topic}.
            As aforementioned, \citet{xu2022hyperminer} use different layers in the hyperbolic embedding space to interpret hierarchical topics.
            \citet{li2022alleviating} employ skip-connections for decoding to alleviate the posterior collapsing issue and propose a policy gradient method for training.
            Recently, \citet{duan2023bayesian} propose to generate different documents for different levels.
            They craft documents with more related words through word similarity matrices for higher levels, and then progressively generate these documents at each level.
            \citet{chen2023nonlinear} utilize a Gaussian mixture prior and nonlinear structural equations to model topic dependencies between hierarchical levels.
            \citet{wu2024traco} propose TraCo.
            It produces sparse and balanced dependencies by modeling them as the transport plan solutions of specially defined optimal transport problems between hierarchical topics.
            They then introduce a context-aware disentangled decoder to decode documents with topics at each level,
            which better distributes different semantic granularity to different levels.

            \paragraph{Non-parametric vs. Parametric}
            As aforementioned, the non-parametric setting allows topic hierarchies to grow dynamically while the parametric setting cannot.
            We must preset the structure of a topic hierarchy in parametric settings.
            However, we still need efforts to determine the hyperparameters of the non-parametric setting, \eg the stick-breaking process prior \citet{chen2021tree}, which control the growing process of topic hierarchies.
            As a result, we conceive that the parametric setting better fits when enough prior knowledge of a dataset is available to determine the topic hierarchy structure ahead.

\begin{figure*}[!t]
    \centering
    \includegraphics[width=0.7\linewidth]{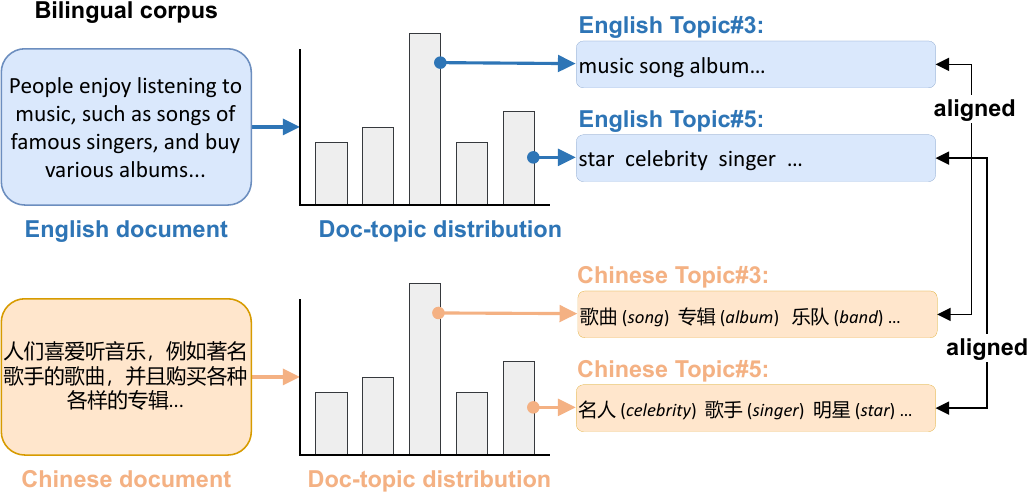}
    \caption{
        Illustration of cross-lingual topic modeling (on English and Chinese documents).
        Corresponding cross-lingual topics are required to be aligned, like English and Chinese Topic\#3, and English and Chinese Topic\#5.
        \emph{Words} in the brackets are the English translations.
    }
    \label{fig_cross-lingual_topic_modeling}
\end{figure*}

        \subsection{Short Text NTMs}
            Researchers apply NTMs to discover topics from short texts.
            Short texts, prevalent on the Internet in various forms such as tweets, comments, and news headlines, serve as a common medium for individuals to express ideas, comments, and opinions.
            However, normal topic models often struggle to effectively handle short texts.
            The principal reason is that topic models depend on the word co-occurrence information to infer latent topics,
            but such information is extremely sparse in short texts due to their limited context.
            This challenge, referred to as \emph{data sparsity} \cite{yan2013biterm,Wu2019}, hinders topic models from discovering high-quality topics and thus has attracted considerable attention in the research community.

            Several studies are proposed to overcome this data sparsity challenge.
            \citet{lin2020copula} use the Archimedean copulas to regularize the discreteness of topic distributions of short texts.
            \citet{Wu2020short} propose NQTM, which quantizes doc-topic distributions of short texts to quantization vectors following the idea of \citet{VandenOord2017}.
            By carefully initializing the quantization vectors, it can produce sharper doc-topic distributions that better fit short texts with limited context.
            They also propose a negative sampling decoder to avoid repetitive topics besides the negative log-likelihood.
            To address the data sparsity issue,
            \citet{wang2021extracting} use word co-occurrence and semantic correlation graphs to enrich the learning signals of short texts.
            \citet{zhao2021entity} propose to incorporate entity vector representations into a NTM for short texts.
            They learn entity vector representations from manually edited knowledge graphs.
            Based on NQTM \cite{Wu2020short},
            \citet{wu2022mitigating} further propose TSCTM, a contrastive learning method according to the topic semantics of short texts,
            which better captures the similarity relations among them.
            This refines the representations of short texts and thus their doc-topic distributions.
            They can also adapt to using data augmentation to further mitigate the data sparsity problem.

        \subsection{Cross-lingual NTMs}
            Cross-lingual NTMs are proposed following cross-lingual topic models \cite{mimno2009polylingual}.
            Cross-lingual topic models aim to discover aligned topics in different languages.
            As exemplified in \Cref{fig_cross-lingual_topic_modeling},
            English and Chinese Topic\#3 both refer to ``music'', and English and Chinese Topic\#5 refer to ``celebrity''.
            In addition if two documents in different languages contain similar latent topics,
            their inferred doc-topic distributions should be similar.
            For instance,
            the doc-topic distributions of the parallel English and Chinese documents in \Cref{fig_cross-lingual_topic_modeling} are similar.
            These aligned cross-lingual topics can reveal commonalities and differences across languages and cultures, which enables cross-lingual text analysis without supervision.

            \citet{Wu2020} propose the first neural cross-lingual topic model, NMTM.
            It transforms the topic-word distribution to the vocabulary space of another language.
            Thus the topic-word distributions of one language can incorporate the semantics of another language, which aligns cross-lingual topics.
            They show that their model outperforms traditional multilingual topic models \cite{shi2016detecting,yuan2018multilingual}.
            Later,
            \citet{wu2023infoctm}
            propose InfoCTM.
            It aligns cross-lingual topics from the perspective of mutual information.
            This can properly align cross-lingual topics and prevent degenerate topic representations.
            To address the low-coverage dictionary issue,
            they also propose a cross-lingual vocabulary linking method that finds more linked words for topic alignment beyond the given dictionary.
            \citet{bianchi2020cross,mueller2021fine}
            directly learn cross-lingual doc-topic distributions with multilingual BERT.
            But we emphasize that they cannot discover aligned cross-lingual topics as required.

        \subsection{Dynamic NTMs}
            Dynamic NTMs are explored following dynamic topic models \cite{blei2006dynamic,wang2008continuous}.
            Previous static topic models implicitly assume that documents are exchangeable.
            However, this assumption is inappropriate since documents are produced sequentially, such as scholarly journals, emails, and news articles.
            As such, dynamic topic models are proposed.
            While topics in previous methods are all static, dynamic topic models allow topics to shift over time to capture the topic evolution in sequential documents.

\begin{figure}[!t]
    \centering
    \includegraphics[width=\linewidth]{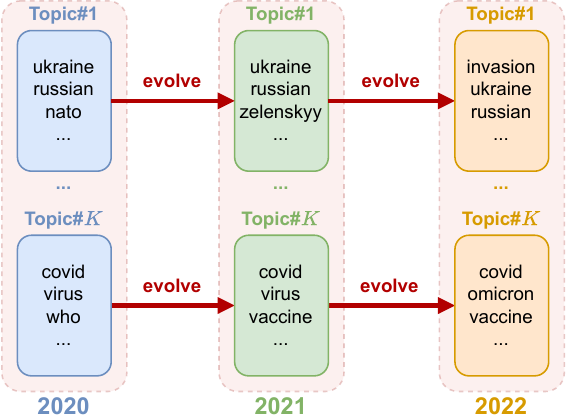}
    \caption{
        Illustration of dynamic topic modeling.
        Topics associated with time slice $t$ evolve from topics associated with slice $t-1$.
        Here Topic\#1 in 2022 evolves from Topic\#1 in 2021.
        It is similar for other topics.
    }
    \label{fig_dynamic_topic_modeling}
\end{figure}

            To be specific, dynamic topic models assume that documents are divided by time slice, for example by year,
            and each time slice has $K$ latent topics.
            The topics associated with slice $t$ evolve from the topics associated with slice $t-1$.
            As the example in \Cref{fig_dynamic_topic_modeling},
            Topic\#1 about Ukraine and Russia evolves from the year 2020 to 2022.
            Due to the emergence of the word ``invasion'', we see Topic\#1 captures the Ukraine-Russia war that exploded in 2022.
            Similarly, Topic\#$K$ about Covid-19 evolves from the year 2020 to 2022 with the explosion of the Omicron variant.
            These topic evolution reveals how topics emerge, grow, and vanish, which has been applied for trend analysis and opinion mining.

            \citet{dieng2019dynamic} first propose a neural dynamic topic model, DETM (Dynamic Embedding Topic Model).
            It uses word and topic embeddings to interpret latent topics following \citet{dieng2020topic}
            and chains topic embeddings at slice $t$ with topic embeddings at slice $t-1$ by Markov chains.
            Besides, it uses a LSTM to learn temporal priors of doc-topic distributions.
            \citet{rahimi2023antm} discover topic evolution by clustering documents but cannot infer doc-topic distributions as required.
            \citet{zhang2022dynamic} focus on the dynamic topics of temporal document networks and incorporate the linking information between documents.
            Following DETM, \citet{miyamoto2023dynamic} propose to employ a self-attention mechanism to model the dependencies among dynamic topics.
            \citet{wu2024dynamic} focus on the unassociated topic and repetitive topic issues.
            Instead of the previous Markov chains fashion,
            they propose CFDTM with a contrastive learning method to resolve these issues and track topic evolution.
            Rather than modeling topic evolution, \citet{cvejoski2023neural} model the activities of topics over time.
            Note that the activities of topics evolve over time but their topics are invariant.
            Thus, this method does \emph{not} precisely adhere to the original definition of dynamic topic modeling.

        \subsection{Correlated NTMs}
            Following the idea of correlated topic modeling \cite{blei2006correlated},
            correlated NTMs have been explored.
            Correlated topic models seek to consider the correlation between latent topics.
            For example, a document about genetics is more likely to be also about disease than x-ray astronomy \cite{blei2006correlated}.
            This leads to better expressiveness than LDA.
            \citet{liu2019neural} follow the VAE-based NTM and use centralized transformation flow to capture topic correlations.
            To effectively infer the transformation flow,
            they present the transformation flow lower bound to regulate the KL divergence term.

        \subsection{Lifelong NTMs}
            Lifelong NTMs are proposed to solve the challenge of data sparsity, similar to short text NTMs but in a continual lifelong learning fashion.
            \citet{gupta2020neural} propose the first lifelong NTM.
            They retain prior knowledge, \ie topics, from document streams and guide topic modeling on sparse datasets with the accumulated knowledge.
            In detail,
            they use topic regularization to transfer topical knowledge from several domains and prevent catastrophic forgetting, and a selective replay strategy to identify relevant historical documents.
            \citet{zhang2022lifelong} propose a lifelong NTM enhanced with a knowledge extractor and adversarial networks.

            Although lifelong and dynamic topic modeling both work on sequential documents,
            we clarify their differences:
            Dynamic topic modeling targets to discover topic evolution, \ie, replacing outdated topics with emergent ones.
            Lifelong topic modeling aims to mitigate the data sparsity issue by accumulating prior topical knowledge, so it needs to restrain from forgetting past knowledge.

\section{Applications of NTMs}
    In this section, we introduce the applications of NTMs, mainly including text analysis, text generation, and content recommendation.

    \subsection{Text Analysis}
        The primary applications of NTMs concentrate on text analysis \cite{boyd2017applications,laureate2023systematic}.
        
        \citet{bai2018neural} apply a NTM to analyze scientific articles.
        They enable a NTM to incorporate the citation graphs of scientific articles by predicting the connections between them.
        Thus their model can also recommend related articles to users.
        \citet{Zeng2018} combine a NTM and a memory network and jointly train them for short text classification.
        Their method classifies short texts and discovers topics from them simultaneously.
        \citet{chaudhary2020topicbert} combine BERT with a NTM, which reduces the operation of self-attention.
        They claim that this can greatly speed up their fine-tuning process and thus reduce CO\textsubscript{2} emission.
        \citet{song2021classification} propose a classification-aware NTM which includes a NTM and a classifier.
        They focus on classifying the disinformation about COVID-19 to help deliver effective public health messages.

        \citet{zeng2019say} apply NTMs to understand the discourse in micro-blog conversations.
        \citet{li2020global} use a dynamic NTM to understand the global impact of COVID-19 and non-pharmacological interventions in different countries and media sources.
        Their discovered dynamic topics help researchers understand the progression of the epidemic.
        \citet{valero2022} propose a short text NTM for podcast short-text metadata.
        \citet{gui2020multi} propose a multitask mutual learning framework for sentiment analysis and topic detection.
        They make the topic-word distributions similar to the word-level attention vectors through mutual learning.
        \citet{avasthi2022topic} use NTMs to mine topics from large-scale scientific and biomedical text corpora.

    \subsection{Text Generation}
        Several studies apply NTMs to text generation tasks.
        Specifically, \citet{tang2019topic} propose a text generation model that learns semantics and structural features through a VAE-based NTM.
        \citet{yang2021topnet} leverage NTMs to alleviate the information sparsity issue in long story generation.
        They map a short text to a low dimensional doc-topic distribution, from which they sample interrelated words as a skeleton.
        With the short text and the skeleton as input, they use a Transformer to generate long stories.
        \citet{nguyen2021enriching} use the doc-topic distributions of NTM to enrich and control the global semantics for text summarization.
        \citet{zhang2022htkg} propose a neural hierarchical topic model to discover hierarchical topics from documents,
        and then generate keyphrases under the hierarchical topic guidance.

    \subsection{Content Recommendation}
        Similar to early work \cite{wang2011collaborative},
        NTMs can cooperate with recommendation systems.
        \citet{esmaeili2019structured} combines a NTM with a recommender system for reviews through a structured auto-encoder.
        \citet{xie2021graph} use a graph NTM for citation recommendation.

\begin{table*}[!t]
    \centering
    \setlength{\tabcolsep}{3mm}
    \renewcommand{\arraystretch}{1.2}
    \resizebox{0.75\linewidth}{!}
    {
        \begin{tabular}{ll}
        \toprule
              & Top words \\
        \midrule
              & \textbf{Trivial Topics} \\
        \textbf{Topic\#1:} & just even like one go really come good away everything \\
        \textbf{Topic\#2:} & abstract accept many long adding bad displayed good additional great \\
        \textbf{Topic\#3:} & fact even really pretty seem seems nothing thing usually often \\
        \midrule
              & \textbf{Repetitive Topics} \\
        \textbf{Topic\#4:} & \uline{sports} scores \uline{games} \uline{soccer} \uline{league} tennis \uline{ncaa} players \uline{football} club \\
        \textbf{Topic\#5:} & \uline{sports} tennis \uline{soccer} \uline{hockey} \uline{games} \uline{football} beach match \uline{players} \uline{ncaa} \\
        \textbf{Topic\#6:} & \uline{sports} match cup \uline{hockey} olympic \uline{football} \uline{players} sport \uline{league} \uline{soccer} \\
        \bottomrule
        \end{tabular}%
    }
    \caption{
        Examples of trivial and repetitive topics.
        Each row represents the top words of a topic.
        Trivial topics include less informative concepts;
        repetitive topics contain repeating words.
        Repetitions are \uline{underlined}.
    }
    \label{tab_quality}%
\end{table*}%

\section{Challenges of NTMs}
    Despite their popularity, NTMs encounter several challenges.
    In this section, we conclude these main challenges as possible future research directions.

    \subsection{Lacking Reliable Evaluation}
        Inheriting from conventional topic models,
        the critical challenge of NTMs primarily lies in the lack of reliable evaluation.
        Current evaluation methods have been developed for years, but they still encounter the following issues.

        \subsubsection{Absence of Standard Evaluation Metrics}
            The topic modeling field lacks standard evaluation metrics.
            Resorting to human judgment provides one effective way to evaluate topic models, such as topic rating and word intrusion tasks \cite{lau2014machine}.
            Unfortunately, its reliance on human raters renders it expensive and time-consuming, limiting its feasibility for wide-scale comparisons.
            Owing to this, researchers generally depend on automatic evaluation metrics, such as the topic coherence and diversity mentioned in \Cref{sec_preliminary_evaluation}.
            However, these automatic metrics encounter the following two problems:
            \begin{itemize}[leftmargin=*]
                \item
                    \textbf{Inconsistent usage of automatic metrics.}
                    The usage and settings of automatic metrics vary across papers and even within a paper.
                    For example, variations include the number of top words, the number of topics, reference corpora, and coherence or diversity metrics.
                    Consequently, the results are often confined to specific studies, impeding the comparability of NTMs across different research papers.
                    Such inconsistencies have led some benchmarking studies to argue that the conventional LDA can still outperform NTMs in certain aspects \cite{doan2021benchmarking,hoyle2022neural}.
                \item
                    \textbf{Questionable agreement between automatic metrics and human judgment.}
                    Some investigations have revealed the discrepancies between the coherence metrics and human evaluation: they find that automatic metrics declare winner models when the corresponding human evaluation does not.
                    This raises concerns that coherence metrics, originally designed for older models, possibly are incompatible with the newer neural topic models \cite{doogan2021topic,hoyle2021is}.
                    We believe similar concerns may extend to diversity metrics: they may also be inconsistent with human assessments.
                    We detail the reasons and offer a heuristic solution in \Cref{sec_TSD}.
            \end{itemize}

            Owing to the above problems, researchers appeal to explore automatic metrics that better approximate the preferences of real-world topic model users \cite{hoyle2021is,stammbach2023re}.
            Thus, proposing standard and practical evaluation metrics is a promising and urgent future research direction for topic modeling.

        \subsubsection{Lacking Standardized Dataset Pre-processing Settings}
            The topic modeling field lacks standardized dataset pre-processing settings for topic model comparisons.
            Researchers routinely pre-process datasets before running topic models, like removing less frequent words and stop words.
            Recent studies find that different dataset pre-processing settings greatly impact topic modeling outcomes, such as the minimum and maximum document frequency, maximum vocabulary size, and stop word sets \cite{Card2018a,wu2023effective}.
            However, these pre-processing settings vary substantially across papers even if they use the same benchmark datasets like 20newsgroup.
            These variations raise questions about the generalization ability of their methods across different pre-processing settings.
            Thus their claimed performance improvements may be untenable.
            In consequence, establishing standardized dataset pre-processing settings emerges as an imperative prerequisite for ensuring reliable and consistent evaluations of topic models.

    \subsection{Low-Quality Topics}
        Regardless of the simplification and popularity of NTMs, the quality of their discovered topics has been questioned from two aspects:
        \begin{itemize}[leftmargin=*]
            \item
                \textbf{Trivial Topics}:
                Discovered topics are trivial with uninformative words. These topics cannot reveal the actual latent semantics of documents.
                As exemplified in \Cref{tab_quality}, the topics include ``even'', ``just'', and ``really''. It is difficult to discern their underlying conceptual semantics.
            \item
                \textbf{Repetitive Topics}:
                Discovered topics are repetitive with the same words, also referred to as the topic collapsing problem.
                As shown in \Cref{tab_quality},
                the topics include the same words like ``sports'', ``games'', and ``soccer''.
                It is hard to distinguish them.
                Apart from that, these repetitive topics imply some semantics are still hidden in documents.
        \end{itemize}
        More disastrously, some NTMs may exhibit triviality and repetitiveness simultaneously in their discovered topics \cite{Wu2020short,wu2023effective}.
        These two kinds of low-quality topics impede the understanding, undermine the interpretability of topic models, and are less beneficial for downstream tasks and applications.
        In consequence, how to effectively and consistently overcome this challenge becomes a necessary and constructive research direction.

    \subsection{Sensitivity to Hyperparameters}
        Another significant challenge of NTMs lies in their sensitivity to hyperparameters.
        Due to the complicated structures, NTMs typically possess more hyperparameters compared to conventional topic models.
        For example, hyperparameters such as dropout probability, batch size, and learning rate assume critical roles in several NTMs \cite{Srivastava2017,Card2018a}.
        Besides, certain NTMs cannot perform well under a large number of topics \cite{Wu2020short}.
        As a result, researchers must meticulously fine-tune these hyperparameters when applying NTMs, especially to new datasets.
        Therefore, the sensitivity of NTMs to hyperparameters curtails the generalization ability of NTMs,
        and this underscores the necessity to mitigate such sensitivity.

\begin{table*}[!t]
    \centering
    \footnotesize
    \renewcommand{\arraystretch}{1.2}
        \begin{tabular}{lrrr}
        \toprule
              \textbf{Top words} & TU    & TD    & \textbf{TSD} \\
        \midrule
        \multicolumn{4}{c}{\textbf{Case 1}} \\
        \textbf{Topic\#1:} \uline{apple} peach grape orange banana & \multirow{3}{*}{0.867} & \multirow{3}{*}{0.733} & \multirow{3}{*}{1.000} \\
        \textbf{Topic\#2:} \uline{apple} company steve \uline{jobs} macintosh &       &       &  \\
        \textbf{Topic\#3:} \uline{jobs} unemployment economy worker salary &       &       &  \\
        \midrule
        \multicolumn{4}{c}{\textbf{Case 2}} \\
        \textbf{Topic\#1:} \uline{apple} peach grape \uline{orange} banana & \multirow{3}{*}{0.800} & \multirow{3}{*}{0.600} & \multirow{3}{*}{0.933} \\
        \textbf{Topic\#2:} \uline{apple} \uline{orange} steve \uline{jobs} macintosh &       &       &  \\
        \textbf{Topic\#3:} \uline{jobs} unemployment economy worker salary &       &       &  \\
        \midrule
        \multicolumn{4}{c}{\textbf{Case 3}} \\
        \textbf{Topic\#1:} \uline{apple} \uline{peach} \uline{grape} \uline{orange} \uline{banana} & \multirow{3}{*}{0.333} & \multirow{3}{*}{0.000} & \multirow{3}{*}{0.000} \\
        \textbf{Topic\#2:} \uline{apple} \uline{peach} \uline{grape} \uline{orange} \uline{banana} &       &       &  \\
        \textbf{Topic\#3:} \uline{apple} \uline{peach} \uline{grape} \uline{orange} \uline{banana} &       &       &  \\
        \bottomrule
        \end{tabular}%
    \caption{
        Comparison of topic diversity metrics under three cases.
        TU (\Cref{eq_TU}) and TD (\Cref{eq_TD}) refer to previous diversity metrics in \Cref{sec_preliminary_evaluation},
        and \textbf{TSD} (\Cref{eq_TSD}) is our proposed new diversity metric.
        Each row represents the top words of a topic.
        Repetitive words are \uline{underlined}.
    }
    \label{tab_diversity_metrics}%
\end{table*}%

\section{Topic Semantic-aware Diversity} \label{sec_TSD}
    In this section, we propose a new diversity metric that considers the semantics of topics when measuring topic diversity.

    \subsection{Problem of Previous Diversity Metrics}
        Previous topic diversity metrics may contradict human judgment.
        These diversity metrics, such as TR \cite{burkhardt2019decoupling}, TU \cite{Nan2019}, and TD \cite{dieng2020topic}, all consider the uniqueness of one top word of topics.
        They believe that diversity is perfect only when each top word is unique.
        However, we argue that this measurement is over-strict since it ignores the fact that different topics may naturally share the same words due to word polysemy.
        As the examples shown in \Cref{tab_diversity_metrics},
        ``apple'' refers to a kind of fruit in Topic\#1 and a technology company in Topic\#2,
        and ``jobs'' refers to Steve Jobs in Topic\#2 or a paid position of employment in Topic\#3.
        These topics imply different conceptual semantics although they include some same words.
        So we conceive their diversity score should be highest.
        But we see that their TU score is only 0.867 and TD is 0.733 in \Cref{tab_diversity_metrics}, which disagrees with our judgment.

\begin{table}[!t]
    \centering
    \setlength{\tabcolsep}{5mm}
    \renewcommand{\arraystretch}{1.1}
    \resizebox{\linewidth}{!}{
        \begin{tabular}{lrrr}
        \toprule
        Datasets & TU    & TD    & \textbf{TSD} \\
        \midrule
        NeurIPS  & 0.970 & 0.970 & 0.984 \\
        ACL   & 0.969 & 0.970 & 0.998 \\
        NYT   & 0.913 & 0.913 & 0.963 \\
        Wikitext103 & 0.973 & 0.973 & 0.999 \\
        \midrule
        \textbf{Average} & 0.957 & 0.957 & \textbf{0.986} \\
        \bottomrule
        \end{tabular}%
    }
    \caption{
        Correlations between topic diversity metrics and human ratings on different datasets.
    }
    \label{tab_diversity_correlation}%
\end{table}%

    \subsection{Topic Semantic-aware Diversity}
        To address this issue,
        we propose \textbf{Topic Semantic-aware Diversity} (\textbf{TSD}), a new metric that measures topic diversity along with word semantics.
        \subsubsection{Definition of Topic Semantic-aware Diversity}
            In detail,
            we compute TSD based on the frequencies of word pairs.
            Given $K$ topics and the top $T$ words of each topic,
            we propose the new TSD as follows:
            \begin{align}
                \mathrm{TSD} \!=\! \frac{2}{K T (T-1)} \!\! \sum_{k=1}^{K}  \sum_{(x_i, x_j) \in t(k)} \!\!\!\!\!\!\! \mathbb{I}( \#(x_i, x_j) ). \label{eq_TSD}
            \end{align}
            Here $\#(x_i, x_j)$ means the number of an unordered word pair $(x_i, x_j)$ in the top words of all $K$ topics.
            $\mathbb{I}(\cdot)$ refers to an indicator function that equals to 1 if $\#(x_i, x_j) = 1$ and equals $0$ otherwise.
            $t(k)$ denotes the top words of $k$-th topic.
            Rather than the uniqueness of one word,
            our TSD measures the uniqueness of word pairs in the top words of topics.
            This is because we know what a word exactly refers to when paired with another one.
            For example, ``apple'' refers to fruit if paired with ``orange'' or ``banana'' and to a company if with ``technology'' or ``company''.
            Note that TSD degrades to TD when measuring the frequency of each word in \Cref{eq_TSD}.

            We exemplify the difference between our TSD with previous diversity metrics.
            \Cref{tab_diversity_metrics} Case 1 shows the TSD score of these three topics is 1.0.
            This is because ``apple'' does \emph{not} co-occur with ``orange'', ``grape'', or ``banana'', and ``jobs'' does \emph{not} co-occur with ``unemployment'', ``economy'', or ``salary'' in Topic\#2.
            Thus TSD considers Topic\#1-3 as different topics regardless of the same words.
            Naturally,
            TSD punishes diversity if the word pairs are repetitive.
            For instance,
            \Cref{tab_diversity_metrics} Case 2 shows the TSD score of the three topics becomes lower since ``apple'' co-occurs with ``orange'' in both Topic\#1 and \#2.
            In the worst situation,
            \Cref{tab_diversity_metrics} Case 3 has all the same topics.
            We see in this case both TD and TSD give 0 for topic diversity.

        \subsubsection{Evaluation Results}
            We conduct experiments to sufficiently compare our proposed topic diversity metric and previous ones.
            In detail,
            we employ a conventional topic model, LDA \cite{blei2003latent} and a neural topic model, NSTM \cite{zhao2020neural} to discover latent topics from real-world datasets. Then we ask human raters to evaluate the diversity among the top words of sampled topics.
            The adopted datasets are listed as follows:
            \begin{inparaenum}[(i)]
                \item
                    NeurIPS~\footnote{\url{https://www.kaggle.com/datasets/benhamner/nips-papers}}, including published papers at the NeurIPS conference from 1987 to 2017.
                \item
                    ACL \cite{bird2008acl}, including research articles between 1973 and 2006 from the ACL Anthology~\footnote{\url{https://aclanthology.org/}}.
                \item
                    NYT~\footnote{\url{https://huggingface.co/datasets/Matthewww/nyt_news}}, including news articles on the New York Times website from 2012 to 2022.
                \item
                    Wikitext103~\footnote{\url{https://www.salesforce.com/products/einstein/ai-research/the-wikitext-dependency-language-modeling-dataset/}} \cite{merity2016pointer}, including Wikipedia articles.
            \end{inparaenum}
            Following \citet{lau2014machine,roder2015exploring}, we compute the Pearson correlation coefficients between the results of these topic diversity metrics and human ratings.

            \Cref{tab_diversity_correlation} shows the correlation results on different datasets and the average correlation.
            We notice that our TSD achieves relatively higher correlation scores with human ratings.
            This is because our TSD metric considers the word semantics as well while measuring topic diversity.
            These empirical results demonstrate that our TSD metric more closely aligns with human judgment concerning the topic diversity evaluation.

\section{Topic Model Toolkits}
    We introduce several topic model toolkits developed by the research community.
    Early popular toolkits include \textsc{MALLET}~\footnote{\url{https://mimno.github.io/Mallet/topics.html}} \cite{mccallum2002mallet}, gensim~\footnote{\url{https://radimrehurek.com/gensim/}} \cite{rehurek2011gensim}, STTM~\footnote{\url{https://github.com/qiang2100/STTM}} \cite{qiang2020short}, ToModAPI~\footnote{\url{https://github.com/D2KLab/ToModAPI}} \cite{lisena2020tomodapi}, and tomotopy~\footnote{\url{https://github.com/bab2min/tomotopy}}.
    However these toolkits often neglect either the implementations of NTMs, dataset pre-processing, or evaluations,
    leaving a gap in meeting practical requirements.
    Recently, \citet{terragni2020octis} propose the \textbf{OCTIS} toolkit~\footnote{\url{https://github.com/MIND-Lab/OCTIS/}}  which includes several NTM methods, evaluations, and Bayesian parameter optimization for research.
    The latest toolkit is \textbf{TopMost}~\footnote{\url{https://github.com/bobxwu/topmost}} \cite{wu2023topmost}.
    While OCTIS only has two NTMs proposed after 2018, TopMost covers more newest released NTMs and a wider range of topic modeling scenarios.
    It also decouples the model implementations and training, which eases the extension of new models.
    These toolkits provide a solid foundation for beginners to explore various topic models and empower users to leverage diverse topic models in their applications.

\section{Conclusion}
    Topic models have been prevalent for decades with diverse applications.
    Recently Neural Topic Models (NTMs) have attracted significant attention due to their flexibility and scalability.
    They stand out by offering advantages such as avoiding the requirement for model-specific derivations and efficiently handling large-scale datasets.
    With the emergence of NTMs, researchers have explored several promising applications for various tasks.

    In this paper, we provide a comprehensive and up-to-date survey of NTMs.
    We introduce the preliminary of topic modeling, including the problem setting, notations, and evaluation methods.
    We review the existing NTM methods that employ different network structures and discuss their applicability to different use case scenarios.
    In addition, we delve into an examination of the popular applications built on NTMs.
    Finally, we identify and discuss the challenges that lie ahead for NTM research in detail.
    We hope this survey can serve as a valuable resource for researchers interested in NTMs and contribute to the advancement of NTM research.

\section*{Acknowledgements}
    We thank all anonymous reviewers for their helpful comments.

\bibliography{lib}

\end{document}